\documentclass[12pt]{article} 

\usepackage[english]{babel}
\usepackage{subfig}
\usepackage{graphicx}
\usepackage{placeins}
\usepackage{soul}
\usepackage{xcolor}
\usepackage{natbib}[nameyear]
\usepackage[letterpaper,top=2cm,bottom=2cm,left=3cm,right=3cm,marginparwidth=1.75cm]{geometry}

\newif\ifnotesw \noteswtrue

\usepackage{amsmath}
\usepackage{graphicx}
    \usepackage{array}

\usepackage[colorlinks=true, allcolors=blue]{hyperref}

\title{Transfer Learning using 66 Diseases for Disease Forecasting Applications}
\author{Lauren J Beesley$^{1,a}$, Alexander C Murph$^a$, Dave Osthus$^a$, Lauren A Castro$^b$\\
\\
$^1$ Corresponding author: lvandervort@lanl.gov\\
$^a$ Statistics, Los Alamos National Laboratory\\
$^b$ Computational Intelligence \& Modeling, Los Alamos National Laboratory}

\usepackage{xcolor}
\colorlet{red}{black} 

\date{}

\begin{document}
\maketitle

\begin{abstract}
Disease forecasting models typically rely on a single data stream, making models brittle when histories are short or noisy. Recent top-performing models have shown that synthesizing multiple reporting systems for the same disease improves performance. Other recent work takes this idea a step further, using transfer learning to train a forecasting model for one disease using data from a different disease. We expand upon each of these approaches greatly, training machine learning models on data that span {\color{red}66} infectious diseases and several data streams. We investigate the value of incorporating different data streams for forecasting {\color{red}20} different disease data streams. We find that incorporating other data streams improves forecasting in the vast majority ({\color{red}84.9}\%) of time series and model structures considered. However, our work highlights that the quality of the added data matters, where adding data extremely different from the target data stream can sometimes degrade forecast performance.  A major contribution of this work is in compiling a publicly-available database of data for use by the infectious disease forecasting community. 
\end{abstract}

\section{Introduction}

Most operational disease forecasts are built around a single data stream, which leaves models data-sparse and brittle when the primary stream is short or noisy.  Recent work has demonstrated that including data from multiple reporting systems can strengthen forecasts in data-sparse scenarios. For instance, the top-performing forecasting model in the 2023/24 FluSight challenge \citep{cdc_flusight_2023_2024_evaluation} used influenza-like illness (ILI) reporting and data on laboratory-confirmed influenza hospitalizations to inform models to forecast hospitalizations reported to the CDC's National Healthcare Safety Network (NHSN) \citep{cdc_nhsn}.  This model, Flusion, used machine learning (ML) to incorporate data from multiple surveillance signals and multiple locations \citep{ray2025}.

The fundamental idea behind Flusion (that data on a single disease, but from multiple reporting systems, can be synthesized to improve forecasting) has been adopted by other recent models. \citet{meyer2025} augmented scant NHSN hospitalization data with ILINet-based reconstructions and prospectively placed 4th (in 2022/23) and 2nd (in 2023/24) in successive FluSight seasons \citep{cdc_flu_burden_2022_2023, cdc_flusight_2023_2024_evaluation}, transferring information across data streams to enable richer ML training.  A similar approach was taken by \citet{benefield2025}, also for the purpose of forecasting influenza hospitalizations.

There are several recent papers that push the cross-data stream learning idea further, incorporating data from distinct pathogens in forecasting.  In \citet{roster2022}, data on Dengue Fever is used to train a forecasting model on Zika, and data from influenza informs a COVID model.  \citet{roster2022} frame this explicitly as transfer learning: using knowledge from one prediction problem to improve another.  Cross-disease transfer learning has been attempted in a handful of other papers to forecast several other diseases \citep{coelho2020, rodriguez2021, chen2022}, although to our knowledge, no prior work uses more than one additional disease during the training step of a ML model. 

Inspired by the recent successes in transfer learning for disease forecasting, this paper pushes this ambition further, using data on {\color{red}66} infectious diseases ({\color{red}76} including disease subtypes, i.e. Hepatitis A and Hepatitis B are subtypes of ``Hepititis") and examining several strategies to determine what information from these diverse data streams can inform infectious disease forecasting models.  Modern predictive models are trained on a large corpus of data and then fine-tuned downstream for specific tasks.  This work compiles and cleans such a corpus, then applies modern predictive models to determine the value of greatly increasing the size of the training data using cross-pathogen and cross-data stream data.  To the authors' knowledge, there have been no modern efforts to incorporate these many data streams/diseases into a single disease forecasting model.  

Like all applications of transfer learning, there is the potential for negative transfer, where incorporating information from other sources leads to a drop in forecasting performance \citep{zhang2023}. Whether including these many data in a disease forecasting model leads to negative transfer is of major concern, and this question is investigated in this paper. One way that this is assessed is by comparing different classes of data that could be fed into the model: 
\begin{enumerate}
    \item \textit{single data stream data}, where only one data source on one disease is used (typical for forecasting applications).  In this paper, a ``data source" refers to a repository from which that data were obtained (e.g., Project Tycho), while a ``data stream" refers to a single disease from a single data source, (e.g., lyme disease from Project Tycho);
    \item \textit{single disease data}, where data across data sources -- but only one disease -- are used \cite[e.g.][]{ray2025, meyer2025, benefield2025};
    \item \textit{single mode of transmission}, where all data across data sources and diseases are used, but only for the same mode of transmission (e.g., respiratory droplets, vector-borne, fecal-oral, etc) \cite[e.g.][]{roster2022, coelho2020, rodriguez2021, chen2022};
    \item \textit{all available data}, across disease, data source, and modes of transmission.
\end{enumerate}
In the forecasting applications in this paper, forecasts are obtained across three types of forecast models fit to each of these classes of training data. 

The organization of the paper is as follows. In Section \ref{sec:data} we discuss the sourcing and cleaning of these data, and give an overview of which pathogens are included.  Then, in Section \ref{sec:methods}, we discuss the forecast models. Lastly, we present results from a comparative study of forecast model performance across {\color{red}20} evaluated data streams (Section \ref{sec:results}). These data streams are diverse, spanning respiratory, sexually-transmitted, fecal-oral, and vector-borne diseases across many distinct data sources and pathogens (Section \ref{sec:results}). In Section \ref{sec:discussion}, we provide a discussion of our results.

\section{Data} \label{sec:data}
\begin{figure}
\centering
\subfloat[Summary of data sources]{\includegraphics[width=0.43\linewidth]{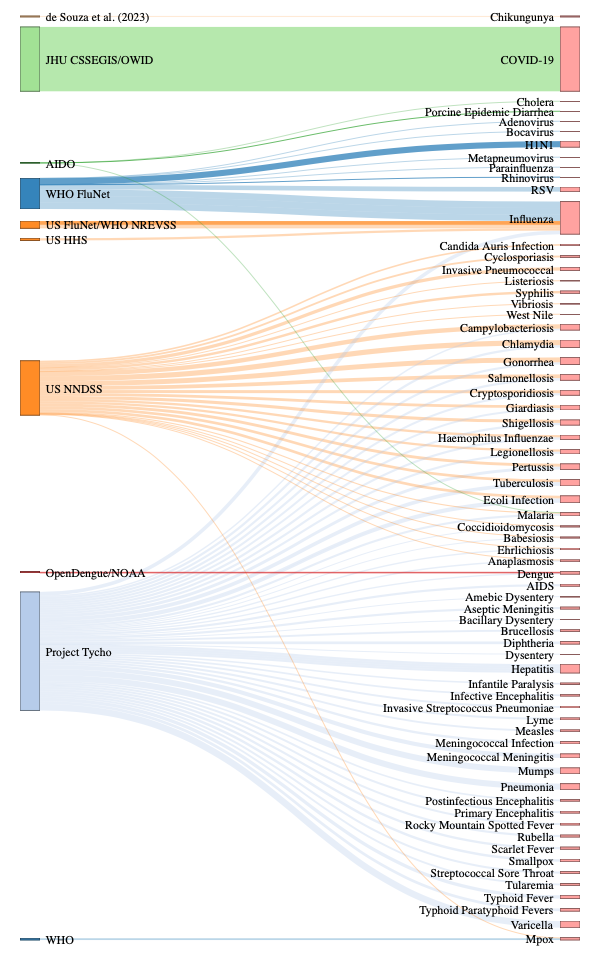}}
\subfloat[Number of non-zero observations]{\includegraphics[width=0.55\linewidth]{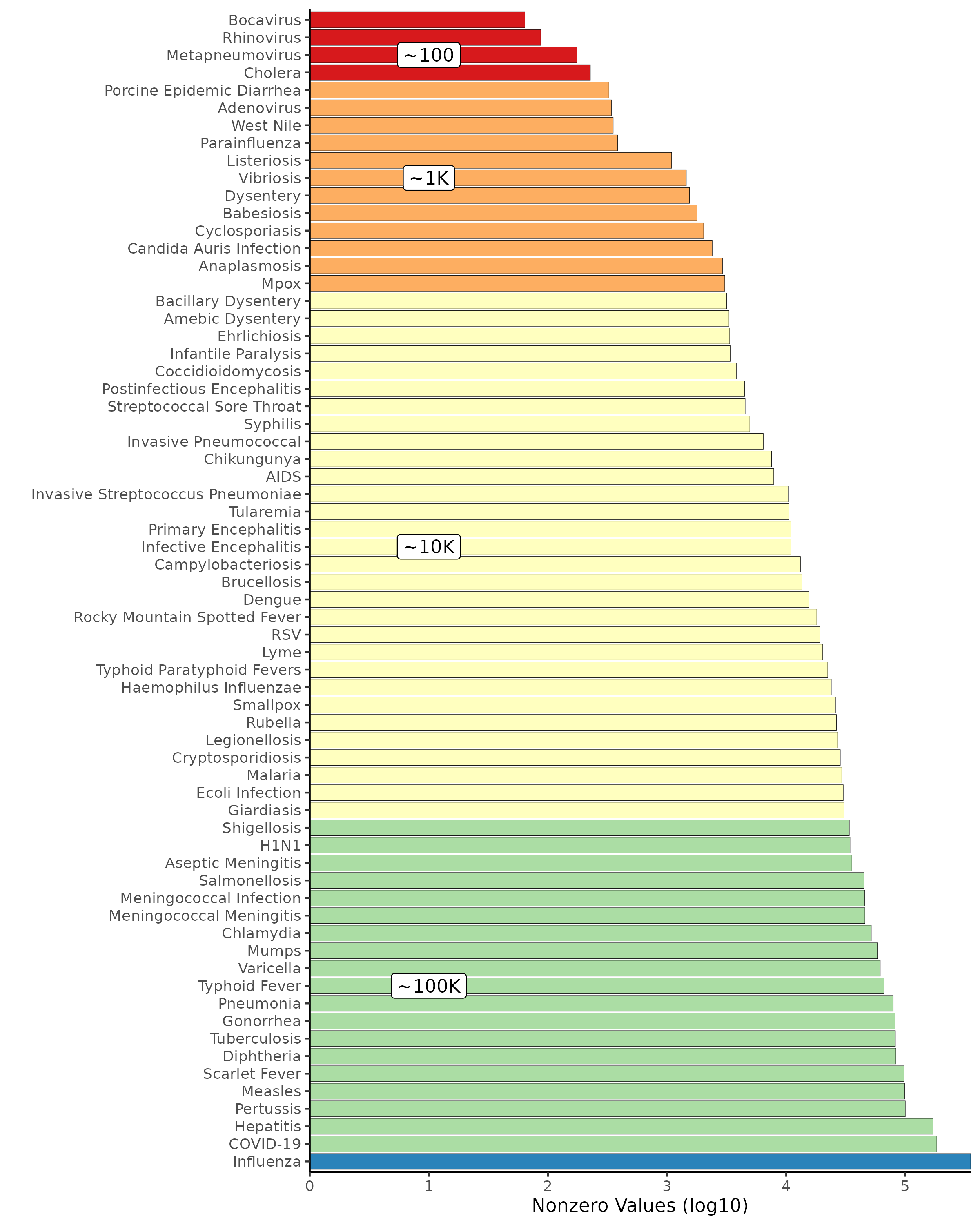}}\\
\caption{\label{fig:data} Visualization of (a) disease data obtained from different data sources and (b) number of non-zero time series observations collected per disease across all sources. In (a), the thickness of each line is proportional to the number of time series available for different locations, and the thickness of the box to the right of each disease is proportional to the total number of time series across sources. Colors on the left-hand side represent different data sources. In (b), rows are ordered according to increasing amounts of data, measured in terms of total number of weeks with non-zero case counts across all available locations. Colors delineate different orders of magnitude for the number of non-zero values.}
\end{figure}

We compiled a large database of infectious disease case, death, and hospitalization data for use in this study. This database spans {\color{red}66} diseases (76 including disease subtypes, e.g., Influenza A and B) across {\color{red}13} data sources as shown in \textbf{Table \ref{tab:data_sources_main}}. For this analysis, we will group some data sources together (OpenDengue and NOAA, JHU CSSEGIS and OWID, US FluNet and US NREVSS) to define {\color{red}10} unique data sources. Defining a data stream as a single disease/disease variation reported from a single data source, these data represent {\color{red}101} distinct data streams. For additional information about data cleaning, licensing (including commercial use of the data), and harmonization, we refer readers to our documented code and dataset available at \url{https://github.com/lanl/precog}. \\
\indent Our analysis focuses on weekly infectious disease time series data. \textbf{Figure \ref{fig:data}} illustrates the amount and variety of time series data collected from each of the {\color{red}10} data sources and {\color{red}66} diseases. In (a), we visualize how the different data sources used in this paper contribute to the data we use to train our ML models. In (b), we visualize how many weeks of non-zero data are available for each of these diseases. By far, the most data available for a disease are for COVID-19 and influenza.  However, across all locations and diseases, there are {over two million} weeks of non-zero case data, spanning as far back as the late 1800s. \textbf{Supp. Figure A.2} visualizes the calendar years with available cases by disease across the various data sources, and \textbf{Supp. Figure A.1} provides more detailed information about the {\color{red}20} data streams used in our evaluation.\\
\indent In addition to taking a more ambitious approach to cross-disease and cross-data stream transfer learning, one contribution of this work is cleaning these data and pulling them into a single repository useful for training future time series forecasters. We hope that future research into transfer learning for disease forecasting will benefit from this large and diverse library of data. Of course, additional data sources or diseases could have been included in our analysis. We narrowed our attention to weekly infectious disease data publicly available online that were somewhat easily accessible, and future work should continue this effort to compile even more complete data repositories for training cross-disease infectious disease forecasting models. \\
\indent Although all the data represented in \textbf{Figure \ref{fig:data}} were used for forecast model training in this work, only a subset of {\color{red}20} diseases and data sources where chosen for forecast model evaluation. We only evaluated time series data meeting the following restrictions: (1) data were available for some time period between 2010 and 2024 and (2) data were available for the disease being forecasted from multiple data sources/data streams. An exception was made for COVID-19 and dengue fever, where in each case the vast majority of the available data for that disease were collected from a single data source. The choice of evaluation data is based on several factors including data availability and quality, as well as whether it was possible to compare data across data streams and mode of transmission, and whether a comparison using all available data was possible.

\begin{table}
\centering
\footnotesize
\begin{tabular}{p{5.5cm} p{3.5cm} p{5cm}}
\hline
\vspace{0.05cm}\\
\textbf{Data Source} &  \textbf{Summary of Data} &   \textbf{Reference} \\
\vspace{0.05cm}\\
\hline
\vspace{0.05cm}\\
Johns Hopkins University Center for Systems Science and Engineering COVID-19 Data Repository (JHU CSSEGIS) & COVID-19 cases/deaths &  \cite{jhucssegis}\\
\vspace{0.05cm}\\
Our World in Data (OWID)&  COVID-19 hospitalizations &   \cite{owid-coronavirus}\\
\vspace{0.05cm}\\
Analytics for Investigation of Disease Outbreaks (AIDO) portal     & Multiple diseases & \cite{aido}\\
\vspace{0.05cm}\\
US National Notifiable Diseases Surveillance System (US NNDSS)    &  Multiple diseases & \cite{nndss}\\
\vspace{0.05cm}\\
OpenDengue   &  dengue fever cases & \cite{opendengue} \\
\vspace{0.05cm}\\
US National Oceanic and Atmospheric Administration (NOAA) dengue fever data   & dengue fever cases & \cite{noaa1} \cite{noaa2} \cite{noaa3}\\
\vspace{0.05cm}\\
Project Tycho    & Multiple diseases &  \cite{projectycho}\\
\vspace{0.05cm}\\
US Health and Human Services (US HHS)& Influenza hospitalizations   & \cite{ray2025}\\
\vspace{0.05cm}\\
World Health Organization FluNet (WHO FluNet)   & Multiple diseases &  \cite{whoflunet}\\
\vspace{0.05cm}\\
US FluNet   & Influenza-like illness & \cite{usflunet}\\
\vspace{0.05cm}\\
US CDC National Respiratory and Enteric Virus Surveillance System (NREVSS)   & Influenza A/B &  \cite{nrevss}\\
\vspace{0.05cm}\\
World Health Organization (WHO)& Mpox cases   & \cite{whompox}\\
\vspace{0.05cm}\\
de Souza Chikungunya study in Brazil & Chikungunya cases & 
\cite{deSouza}\\
\vspace{0.05cm}\\
\hline
\end{tabular}
\caption{\label{tab:data_sources_main}Overview of different data sources included in our analysis. }
\end{table}

\FloatBarrier
\section{Methods} \label{sec:methods}
\begin{figure}[h]
\centering
\includegraphics[height=2.5in]{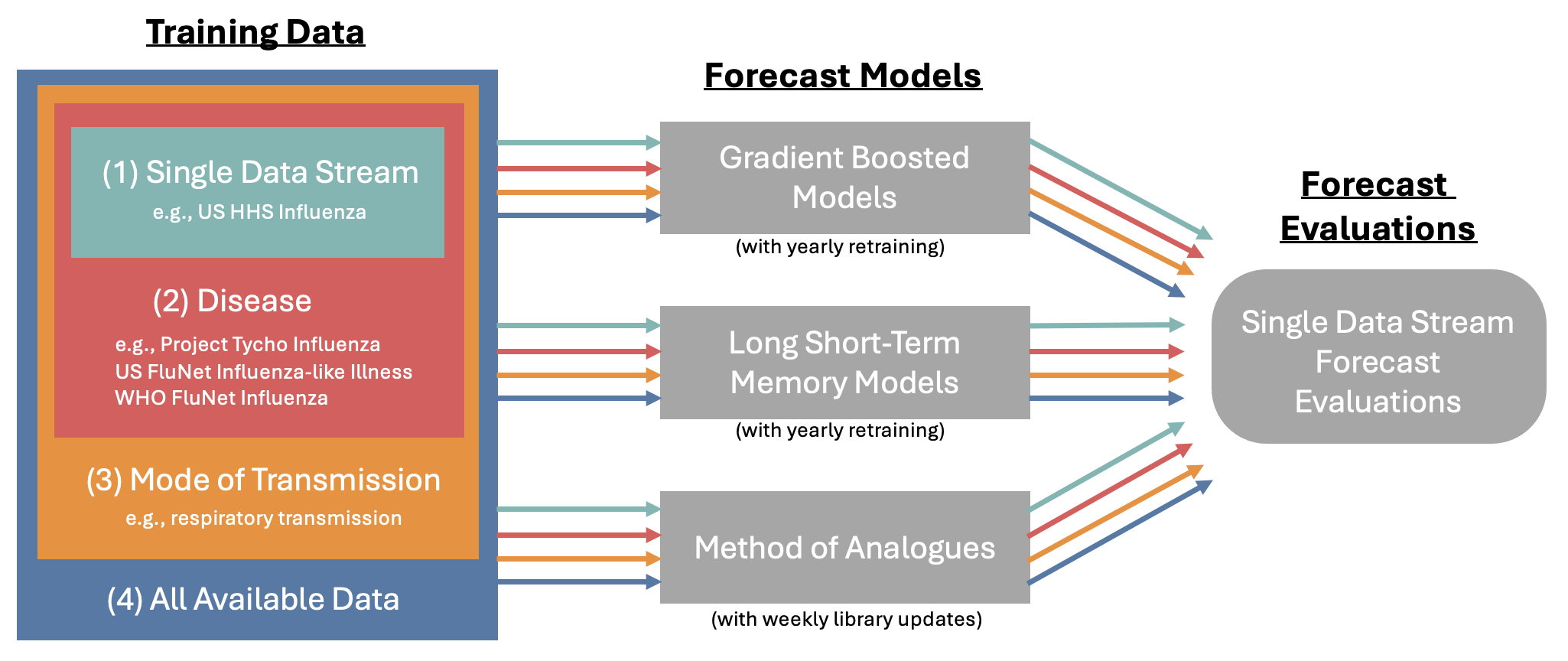}
\caption{\label{fig:methods_diagram} An example of a schematic of various forecast model evaluations and training data comparisons. In this example, three models to predict US Health and Human Services Influenza Hospitalizations are each built from four different data sources. }
\end{figure}

In this section, we describe our approach to time series forecasting using three different forecast model approaches: gradient boosted models (GBM), long short-term memory models (LSTM), and the method of analogues (MOA). Within each forecast model class and data stream evaluated, our goal was to compare forecasting performance obtained using different subsets of the available data for training. \\
\indent Below, we describe these three forecast modeling approaches in general. We also detail our forecast evaluation methods, which focused on quantifying and comparing forecast accuracy and probabilistic forecast calibration within each of the three modeling classes. We provide an overall schematic of our methodology in \textbf{Figure \ref{fig:methods_diagram}}. For each data stream evaluated, we fit each of the three forecasters (GBM, LSTM, MOA) for up to four different formulations of training data (single data stream, all data for the given disease, all data from the same mode of transmission, and all available infectious disease data--possibly with some sub-sampling). Each of the GBM and LSTM models were retrained yearly (e.g., at the start of 2010 to forecast 2010, at the start of 2011 to forecast 2011), while the MOA forecasts were obtained by updating the training data library each week on a rolling basis. Probabilistic uncertainties were obtained for GBM and LSTM models using quantile-specific models with a pinball loss (described below), and uncertainties for MOA forecasts were obtained negative binomial modeling as in \citet{Murph2025}. 

\subsection{Forecast Models}
Here, we describe three different forecast modeling approaches evaluated in this work. See the documented code at \url{https://github.com/lanl/precog/tree/main/crossdisease} for additional details on each model implementation.

\subsubsection*{Gradient Boosted Model}
We first consider forecasts generated using gradient boosted modeling (GBM), a machine learning method that can be viewed as a boosted random forest constructed to capture gradients of the predictive surface. We estimate GBM model parameters using LightGBM, a faster approximation of the popular XGBoost algorithm, as implemented in the \textit{R} package \textit{lightgbm}. \\
\indent This model generates forecasts by learning the relationship between pre-calculated, custom features of the time series being forecasted and the subsequent future time series observations. These time series features could capture the local-in-time trends of the time series, the strength of long-term seasonality, whether the time series is likely near a local maximum/minimum, etc. \\
\indent In our implementation of this model, we defined a wide variety of time series features that could potentially be useful for infectious disease forecasting. There are many existing packages that can calculate standard time series features, but we instead chose to construct our own with the goal of identifying features that helped capture behavior commonly seen in infectious disease case forecasting (as opposed to, say, sales figures). The time series features used in our analysis are detailed in \textbf{Supp. Table A.1}. These features expanded upon those developed for infectious disease time series in \citet{Murph2026}. Naturally, we could imagine many other time series features that could be useful in such a model, and our goal was not to define the features with the highest predictive accuracy possible; rather, we wanted to construct a reasonable model of sufficient quality to allow us to explore the relative performance of the model forecasts obtained with different sets of training data and for different forecast targets (e.g., dengue fever vs COVID-19). \\
\indent Each included time series $ts(1), \hdots, ts(t)$ was used to construct many rows of the GBM training data, where separate rows were generated for each sub-time series $ts(1), \hdots, ts(u)$ for $u = 11, \hdots, t$. Features in each row of the training dataset were calculated on scaled time series that had passed through an outlier filter. Outliers were identified separately for each row of the training dataset using a custom outlier rule as described in \textbf{Supp. Section A}. \\
\indent GBM model parameters were estimated using a pinball loss objective function (i.e., using quantile regression), with separately-trained models for each $\alpha \in (0.025, 0.1, 0.25, 0.5, 0.75, 0.9, 0.975)$. These models allowed us to obtain probabilistic forecasts corresponding to 50\%, 80\%, and 95\% predictive intervals.\\
\indent We view GBMs as representing a large class of traditional machine learning predictive models, where predictions are functions of user-defined, bespoke features abstracted from the available data. These types of models have been popularized for use with time series forecasting based on their strong performance in several recent forecasting challenges \citep{ray2025, M4, M5}.

\subsubsection*{Long Short-Term Memory Model}
We also obtain forecasts using a Long Short-Term Memory Model (LSTM) as implemented by the \textit{keras} package in \textit{R}. We train the model to provide one-week-ahead forecasts based on the last 52 weeks of data (if available). Forecasts for more than one week ahead are obtained by chaining prediction calls together (e.g., by treating the one-week-ahead forecasts as observed data for generating two-week-ahead forecasts). For scenarios with less than 52 weeks of time series data available, missing observations are replaced with -1 and masked out within the fitting process. Similar to the GBM training data construction, each included time series $ts(1), \hdots, ts(t)$ was used to construct many rows of the LSTM training data using a rolling last observation approach.   \\
\indent The LSTM model architecture relied on software defaults for all hyperparameters (e.g., convergence tolerances). The full model structure featured a masking layer to handle missing observations, two LSTM layers, two dropout layers to help avoid overfitting, and a final dense layer. This model structure was not optimized for these data; rather, we posited a reasonable model structure and did not adjust the model structure to optimize forecasting performance in any way. Parameters were estimated using a pinball loss objective function, with separately-trained models for each $\alpha \in (0.025, 0.1, 0.25, 0.5, 0.75, 0.9, 0.975)$. \\
\indent We view the LSTM model architecture as representing a wider class of modeling approaches that leverage machine learning but do not rely on custom, pre-calculated time series features for forecasting. This model has the potential to learn from both short term and longer-term (in our implementation, up to a year) behavior.

\subsubsection*{Method of Analogues}
The final forecasting approach we consider is a variation of the method of analogues (MOA). This non-parametric forecasting approach works by first considering the last $k$ (in our implementation, $k=5$) time series observations, denoted $x_{test}$. We will call this a time series ``snippet."  Prior to forecasting, we define a large library $X$ containing length $k$ time series snippets based on the defined training data, where each snippet is also paired with a corresponding vector representing the following $H=4$ observations. We denote the paired library of future observations $y$.  \\
\indent Using the library of training data snippets of length $k$, we identify a ``neighborhood" of library snippets in $X$ with the $L$ smallest L1-norm values relative to $x_{test}$. We define our forecast by taking the median across the subset of $y$ corresponding to the identified neighborhood of $X$. Based on prior work in \citet{Murph2025}, we set $L = 4,422$ (or 10\% of the number of library snippets, whichever is smaller), but different values could have been chosen.  \\
\indent For each forecast date, we also obtained 50\%, 80\%, and 95\% prediction intervals using the method in \citet{Murph2025}. This method trains a negative binomial distribution on an online history of forecast residuals over time.  Typically, a separate Negative Binomial distribution is fit for different forecast horizons, using the history of residuals at that horizon. \\
\indent The MOA forecasting approach differs from both the GBM and LSTM in that it relies solely on very short-term time series relationships. Unlike GBM and LSTM, it may be viewed as a more ``traditional" time series forecasting method. In our implementation, however, we deviate from a traditional use case by sometimes defining the library of training snippets based on a disease data from many different contexts rather than the single time series being forecasted.

\subsection{Time Series Scaling}
When training a forecast model using time series with large differences in magnitude (e.g., proportions vs case counts), a natural question is whether data with different magnitudes should be rescaled/standardized to a common scale. Even when modeling a single data stream measured for different locations, time series standardization is generally useful since it normalizes each training data in terms of magnitude and relative scale. \\
\indent In this work, we train and deploy our forecast models using rescaled time series obtained by dividing the time series observations by the last observed value (or the last, smoothed value obtained after applying a Generalized Additive Model over the raw observations as in the case of GBM). In particular, each included time series $ts(1), \hdots, ts(t)$ was used to construct many rows of the training dataset, where separate rows were generated by each sub-time series $ts(1), \hdots, ts(u)$ for $u = 11, \hdots, t$. In practice, we rescale the time series used to construct each row by $ts(u)$ (or by its smoothed analogue). If $ts(u)$ is zero, no scaling is applied. \\
\indent We chose to implement rescaling by the last time series observation, but there are many other time series scaling approaches we could have used. For example, a popular rescaling approach transforms the time series by subtracting the mean and dividing by the standard deviation. We do not provide a comparison of different time series rescaling approaches here. See \citet{lima2023, tawakuli2025} for recent reviews of time series rescaling methods for machine learning modeling. Since our goal was to compare forecast model performance \textit{across different sets of training data}, and we do not expect the qualitative results presented in this work to change much as a function of time series rescaling methodology, assuming some sort of reasonable rescaling is implemented.

\subsection{Forecast Model Evaluation} \label{subsec:forcast_evaluation}
We evaluate each set of forecasting models to forecast {\color{red}20} different data streams with data available between 2010 and 2024 (or a subset of these years) and a total of up to {\color{red}2,926} forecasted time series (this varied between methods). The full set of available training data included results from {\color{red}6,875} time series across {\color{red}76} data streams. \\
\indent The exact evaluation time period varies between data streams and methods. For MOA, the evaluation period for a given data stream starts on date A and continues through the last date with available data, with date A defined as either the first week of 2010 or the first week with at least 11 weeks of prior data available. For GBM and LSTM, the evaluation period for a given data stream starts when roughly one year of data is available for that data stream. The snippet library used for MOA forecasting is updated as new data is becomes available during the evaluation period. GBM and LSTM forecasting models are retrained yearly (as opposed to weekly), where the forecasts during a given year are provided using models trained on data available at the start of that year.\\
\indent For each model type (MOA, GBM, LSTM), we generate weekly one- to four-week-ahead forecasts for each week during the evaluation period.  Forecasts are generated using four different sets of training data, as visualized in \textbf{Figure \ref{fig:methods_diagram}}: (4) all available data streams/diseases, (3) data for diseases with the same mode of transmission (respiratory, sexual, fecal-oral, or vector-borne), (2) all data streams available for a particular disease, and (1) data from a single data stream and disease only. Datasets (4) to (1) represent increasingly narrow sets of data included for training. An exception to strict nesting was made for datasets (3) and (4), as the training data rows were sub-sampled by 50\% for (4) due to the large volume of data streams and diseases included. \\
\indent For each data stream and analysis type, we summarized the forecast accuracy across all weeks in the evaluation period using the forecast mean absolute error (MAE), defined as follows:
\begin{align*}
MAE = \frac{1}{H (T-t^*)}\sum_{t=t^*}^T \sum_{h=1}^H \vert f(t+h \vert t) - ts(t+h) \vert, 
\end{align*}
where $h$ denotes the forecast horizon in weeks, $t$ denotes the last week of available data, $ts(t+h)$ is the true time series value being forecasted, and $f(t+h \vert t)$ is the $h-$week forecast using data available up to week $t$.  The value $H$ indicates the maximum horizon considered for forecasting ($H = 4$ in all applications in this paper) and $T$ indicates the final time point used for training. The value $t^*$ differs by method and time series but represents the first time in the evaluation period in which at least 11 prior observations were available for MOA and roughly a year of observations were available for GBM and LSTM. \\
\indent For each data stream, analysis type, and evaluation week, we also generated 50\%, 80\%, and 95\% prediction intervals. We evaluated these probabilistic forecasts using two common metrics: average weighted interval score (WIS) and coverage of 95\% prediction intervals. We define the coverage of the 95\% prediction intervals as
\begin{align*}
Coverage = \frac{1}{H (T-t^*)}\sum_{t=t^*}^T \sum_{h=1}^H I(ts(t+h) \in [q_{0.025}(t+h \vert t), q_{0.975}(t+h \vert t)]), 
\end{align*}
where $q_{u}(t+h \vert t)$ is the $u^{th}$ quantile of the predictive distribution for time $t+h$ using the data available up to time $t$, and $I()$ is an indicator function. The value $t^*$ differs by method and time series as above, but MOA uncertainty evaluation required at least 20 prior observations as opposed to 10 required for MAE evaluation. The average weighted interval score \citep{bracher2021evaluating} is defined as
\begin{align*}
WIS = \frac{1}{K + 0.5}\frac{1}{H (T-t^*)}\sum_{t=t^*}^T \sum_{h=1}^H  \left\{ \frac{\vert f(t+h \vert t) - ts(t+h) \vert}{2} + \sum_{k=1}^K  \frac{\alpha_k}{2} IS_{\alpha_k}(F, ts(t+h))  \right\},
\end{align*}
where
\begin{align*}
IS_{\alpha_k}(F,ts(t+h)) &= q_{1-\alpha_k/2}(t+h \vert t) - q_{\alpha_k/2}(t+h \vert t) \\
&+ \frac{2}{\alpha_k} [q_{\alpha_k/2}(t+h \vert t)-ts(t+h)] \times \mathbf{I}( ts(t+h)< q_{\alpha_k/2}(t+h \vert t)) \\ 
&+ \frac{2}{\alpha_k} [ts(t+h)-q_{1-\alpha_k/2}(t+h \vert t)] \times \mathbf{I}( ts(t+h)> q_{1-\alpha_k/2}(t+h \vert t)) .
\end{align*}
Since we calculate the 50\%, 80\%, and 95\% prediction intervals, we set $K=3$ and $\alpha = (0.5, 0.2, 0.05)$.


\section{Results}\label{sec:results}

\subsection{Analysis of forecasting performance across different training datasets and forecast models}

Across forecast models, evaluation data streams, and training datasets, we performed {\color{red}21,448,672} individual forecasts according to the process outlined in Section \ref{subsec:forcast_evaluation}.  Across all time series forecasted and across all three models, forecasting using only the single data stream being evaluated rarely produced the best MAE (only {\color{red}15.1}\% of the time) relative to forecasting using an expanded set of training data. Single data stream forecasters were outperformed {\color{red}61}\%, {\color{red}78}\%, and {\color{red}72}\% of the time by forecasters trained using all disease, all mode of transmission, and all available data models, respectively. Excluding MOA, these percentages rose to {\color{red}66}\%, {\color{red}84}\%, and {\color{red}81}\%. In aggregate, this highlights the potential for positive transfer learning when incorporating other data streams. However, it also demonstrates the potential for negative transfer, since using all the available data sometimes resulted in worse performance than using only data with a shared mode of transmission.

\begin{figure}
\centering
\caption{\label{fig:mae_comparison} Forecast accuracy obtained using different training datasets. Forecast accuracy is measured by mean absolute error (MAE), relative to a model estimated using only historical data from the data stream being forecasted. Panel (a) provides boxplots of MAE ratio across all time series evaluated and panel (b) provides average results for selected data streams. Panel (c) summarizes the average coefficient of variation, dataset size, and sample entropy for the different training datasets in (b). Points in (a) correspond to distinct time series. Points in (b) and (c) correspond to averages across time series for a given data stream. The coefficient of variation is calculated for each sliding window training data point using the prior 10 weeks of available data and then averaged across all rows of the training data. The sample entropy is calculated for each training data time series and then averaged across all time series in the training data.   }
\subfloat[Boxplots of MAE ratio across evaluated time series]{\includegraphics[width=1\linewidth,trim={0cm 0cm 0cm 0.6cm},clip]{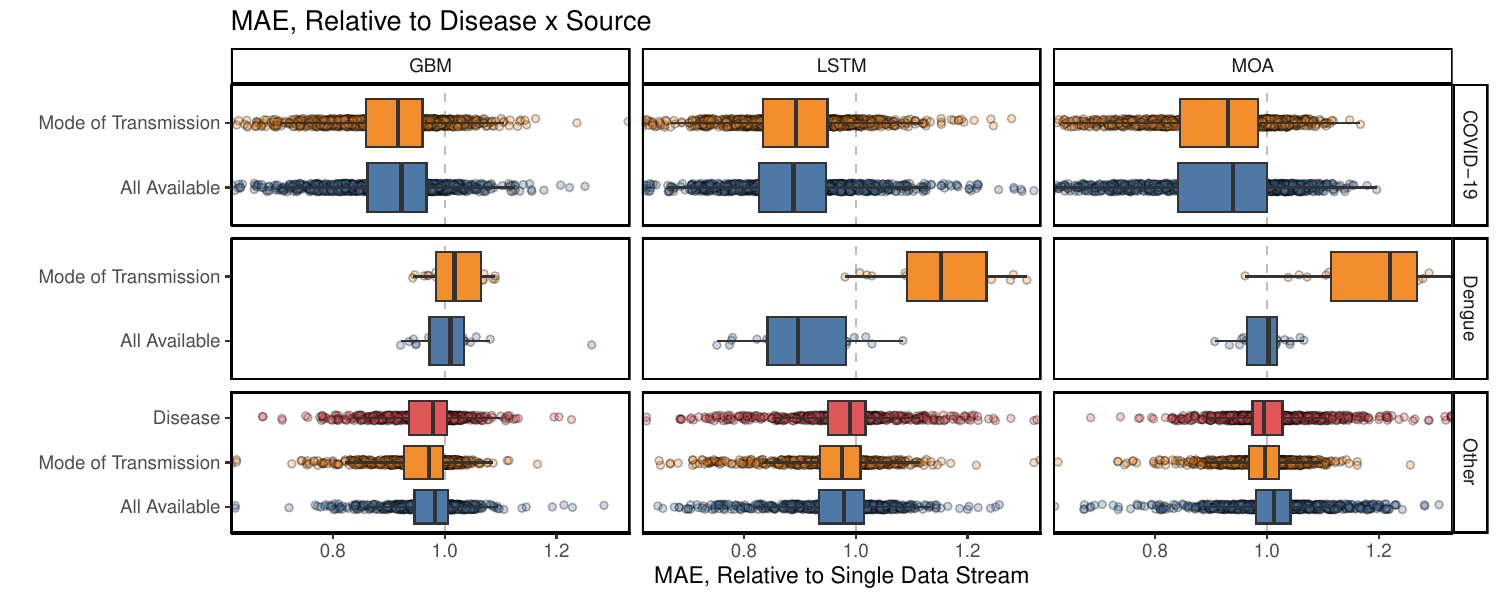}}\\
\subfloat[Overall MAE ratios by disease for selected data streams]{\includegraphics[width=1\linewidth,trim={0cm 0 0cm 1cm},clip]{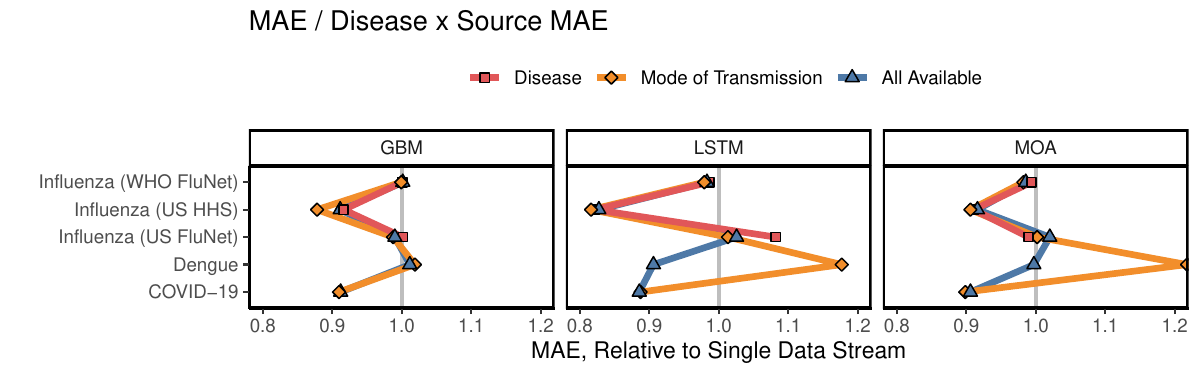}}\\
\subfloat[Summaries of time series training data for selected data streams]{\includegraphics[width=1\linewidth,trim={0cm 0 0cm 1cm},clip]{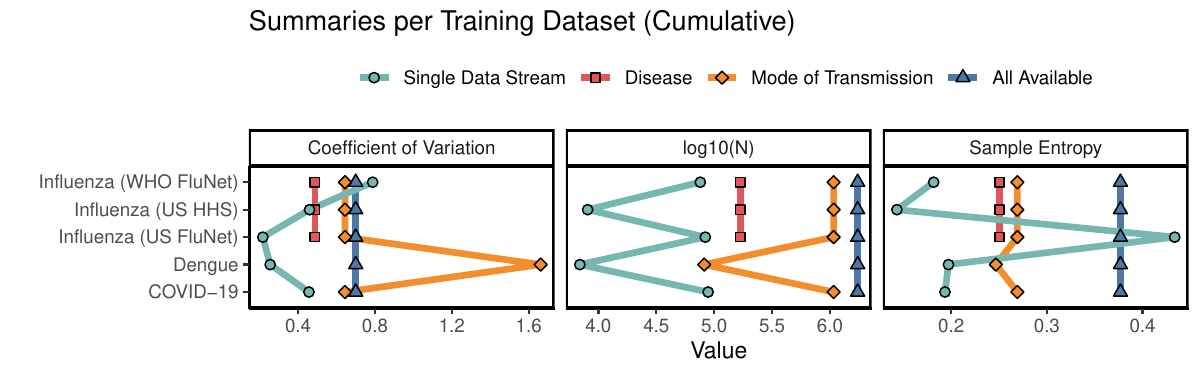}}
\end{figure}

In \textbf{Figure \ref{fig:mae_comparison}a}, we visualize the distributions of MAE across these forecasts using boxplots, where each point in \textbf{Figure \ref{fig:mae_comparison}a} represents a unique time series being forecasted. Similar results for weighted interval score and 95\% prediction interval coverage are provided in \textbf{Supp. Figure B.3}. To allow comparison of MAEs across times series with vastly different case count magnitudes, the MAE values in \textbf{Figure \ref{fig:mae_comparison}} are normalized according to the single data stream (i.e., US HHS influenza hospitalizations) MAE for each time series. Values to the left of the vertical dashed lines correspond to time series in which the evaluated forecaster outperformed the forecaster trained on the single data stream on average in terms of MAE. Since COVID-19 forecasts accounted for over half of the time series being forecasted, we separated out the forecasting performance between COVID-19 time series and all other time series.  We also separated out OpenDengue dengue fever evaluations as this data stream exhibited unusual behavior. 

Across all three models, there is a noticeable improvement in relative MAE for COVID-19 forecasts once all data with the same Mode of Transmission are added. Forecast results using all available data (including other modes of transmission) for training were similar to training using only respiratory diseases for COVID-19.  

For forecasting dengue fever, the forecasters trained using all available vector-borne disease data tended to perform \textit{worse} than those forecasters trained just using OpenDengue dengue fever data. We will discuss this issue more later. Dengue fever forecast performance using all available data varied by model, with GBM and MOA forecast performance similar on average to forecasting using only the OpenDengue dengue fever data. For LSTM, however, there was a meaningful improvement in forecast performance for dengue fever from using all available time series data. 

For forecasts of the other diseases, there is some improvement in the overall performance of the GBM and LSTM models when adding other data streams and other-disease data within the same mode of transmission in sequence, relative to forecasting using the single data stream. For MOA, there is not a clear improvement seen by adding other-disease data with the same mode of transmission. For all three models, we see similar or worse performance when we train using all of the available infectious disease data relative to training using all data with the same mode of transmission. We hypothesize this decrease in performance may have to do with a lack of similarity between the time series being evaluated and the diseases added to the ``all available" training dataset in terms of time series noise, where the methods evaluated may struggle to correctly discriminate between low-noise and high-noise data. Since our models all relied on last-observation scaling for forecasting, the models may not have had enough information to correctly distinguish between low counts (and therefore high variability) settings vs high counts and low vs high noise. See \textbf{Supp. Section B} for a more detailed discussion of this issue.  

We highlight the comparative analysis from \textbf{Figure \ref{fig:mae_comparison}a} for several data streams in \textbf{Figure \ref{fig:mae_comparison}b}. \textbf{Supp. Figures B.1 and B.2} compare forecast MAE as in \textbf{Figure \ref{fig:mae_comparison}} but stratifying results by mode of transmission and by individual data stream, respectively. A detailed discussion of these results is provided in \textbf{Supp. Section B}. Dramatic improvements over training using the single data stream are evident for both COVID-19 and US HHS influenza hospitalizations (so, the plotted points tend to be to the left of the vertical gray line). As an initial observation, the Flusion model from \citet{ray2025} performed forecasts on influenza (US HHS) and discovered that including influenza data across several reporting sources improved performance. These results are corroborated here; across all three models, we see meaningfully improved forecast performance by including other data streams in training relative to training using only the US HHS influenza data alone. In fact, all models saw an additional boost in US HHS influenza forecast performance by including other respiratory disease data in training. There is comparatively little improvement over single data stream forecasts for WHO and US FluNet data. This is likely due to the large volume of WHO and US FluNet data, where single data stream forecasts already rely on a large amount of available training data. 

As mentioned earlier, OpenDengue dengue fever comparative performance is surprising, where training using all vector-borne disease data results in markedly worse performance than just using the OpenDengue data alone or using all the available data. In \textbf{Figure \ref{fig:mae_comparison}c}, we explore this issue in more detail. For each set of training data for the various models in \textbf{Figure \ref{fig:mae_comparison}b}, we summarize (1) the average coefficient of variation in the recent past (a metric of the local-in-time signal to noise ratio, where larger values indicate larger noise relative to the signal), the (2) total amount of training data, and (3) the average sample entropy (a metric of how unpredictable/complex the time series is, where larger values indicate a harder-to-forecast time series).  We chose these specific summaries because of their potential to explain significant differences in how well/poorly each model can learn using the possible training data subsets. Consider the entries for dengue fever. We see a huge difference in the average coefficient of variation between the training data using only OpenDengue data (single data stream) and all vector-borne disease data (mode of transmission). One plausible explanation for the degrading forecast performance for a model trained using all vector-borne disease data is that the data being added are simply too different from the dengue fever data to be useful for forecasting. This would not be a problem if our forecaster was able to identify that these added time series are too different and downweight their contribution. Our GBM model explicitly accounts for the estimated coefficient of variation, providing a route for discriminating between these different scenarios, but the LSTM and MOA models do not directly have a way to capture these differences. This could explain why the drop in forecasting performance is minor for the GBM in comparison to the LSTM and MOA models in \textbf{Figure \ref{fig:mae_comparison}a} and \textbf{Figure \ref{fig:mae_comparison}b}.  A further analysis using these summaries on diseases other than dengue is available in the \textbf{Supp. Section B}.

\begin{figure}
\centering
\caption{\label{fig:mae_explore} Relationships between forecast performance, forecastability, and amount of single data stream data. MAE ratio is related to (a) single data stream training data size as measured by the number of MOA embeddings and (b) time series forecastability as measured by sample entropy. Panel (a) points represent median MAE ratios across all time series and the three time series forecasting models (MOA, GBM, LSTM) for one of the 20 evaluated data streams. Panel (b) points represent individual time series observed for each of the 20 data streams. MAE ratio is calculated as the mean absolute prediction error using the corresponding training dataset, relative to the error from training using data from the single data stream. Panel (a) excludes data streams with more than 20,000 single data stream embeddings (e.g., COVID-19).   }
\subfloat[MAE ratio as a function of single data stream training data size ]{\includegraphics[width=0.7\linewidth,trim={0cm 0cm 0cm 0.5cm},clip]{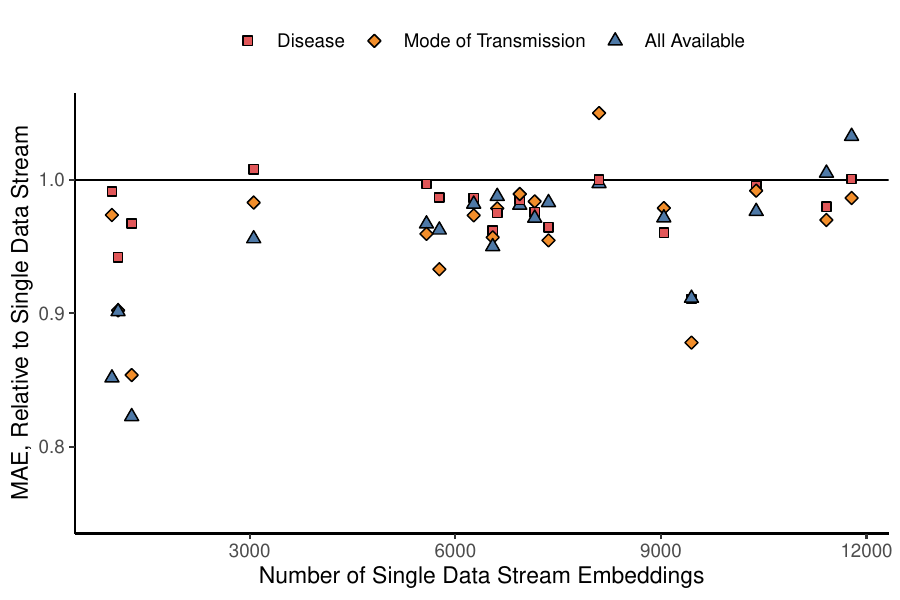}}\\
\subfloat[MAE ratio as a function of time series sample entropy (i.e., forecastability)]{\includegraphics[width=1\linewidth,trim={0cm 0cm 0cm 0.7cm},clip]{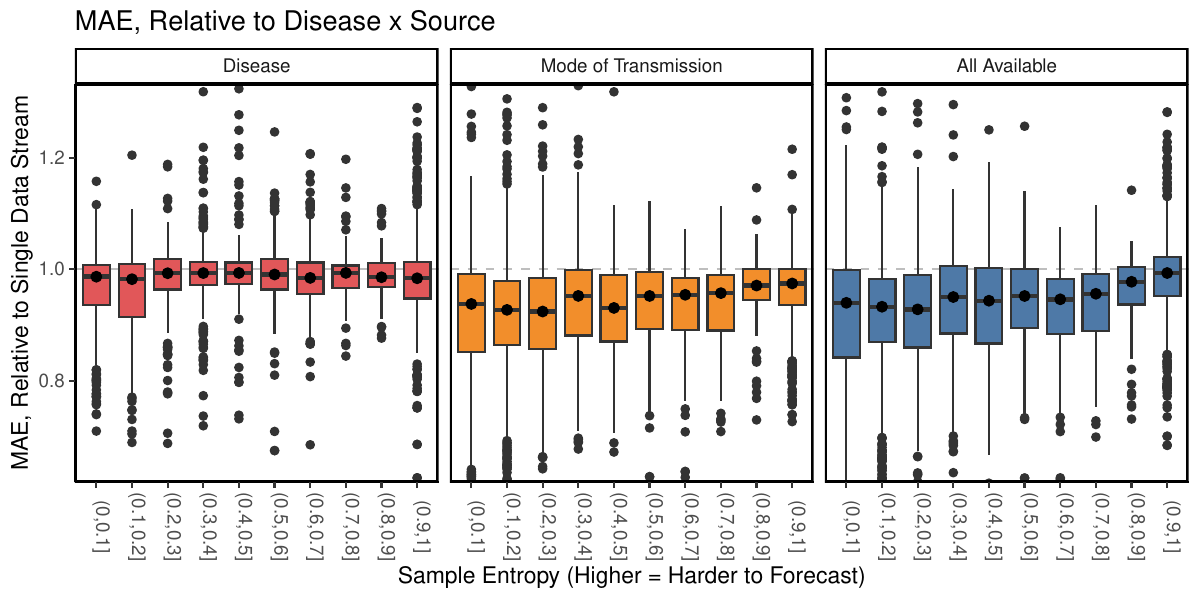}}\\
\end{figure}

In \textbf{Figure \ref{fig:mae_explore}}, we explore how the potential to improve single data stream-based forecast model training relates to (a) the amount of single data stream data and (b) the sample entropy (i.e., forecastability, where higher values of sample entropy indicate harder-to-forecast time series). Perhaps unsurprisingly, we tend to see the largest improvements from including other data streams when the data stream being forecasted has less available historical data. Sample size does not fully explain the relative performance, as large single data stream sample size settings sometimes still had low relative MAE (e.g., the points around 9500 in (a)). 

We also observe a relationship between the benefit of training using more training data (all disease $<$ all mode of transmission $<$ all available) and the sample entropy, where low-entropy time series (i.e., time series that are easier to forecast) appear to benefit more from incorporating more sources of data on average than time series that are harder to forecast. This may be because there is just not much potentially exploitable information contained in these time series to inform forecasting or that the single data stream data provide the most accurate information about the unique features of those harder-to-forecast data. 

\subsection{Information from several diseases contribute to COVID-19 forecasts}
\begin{figure}
\centering
\caption{\label{fig:covid_casestudy} Composition of identified neighborhoods used for COVID-19 MOA forecasting, with the snippet library defined using all available data. Panel (a) shows the proportion of MOA neighborhood snippets by disease and panel (b) shows these proportions scaled relative to the composition of snippets in the entire forecasting library. Panels (a) and (b) provide results aggregated across all COVID-19 time series evaluated, included cases, deaths, and hospitalizations. Panel (c) shows neighborhood composition probabilities for forecasting US overall COVID-19 cases by week, where points represent scaled COVID-19 cases per week and where COVID-19 and influenza neighborhood proportions are stacked.  }
\subfloat[Neighborhood composition]{\includegraphics[width=0.45\linewidth]{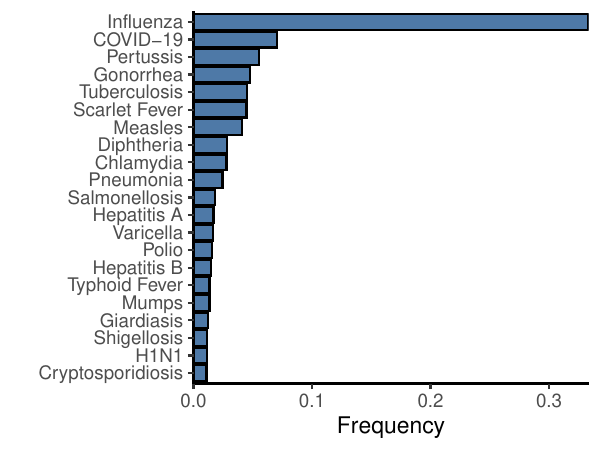}}
\subfloat[Relative to library composition]{\includegraphics[width=0.45\linewidth]{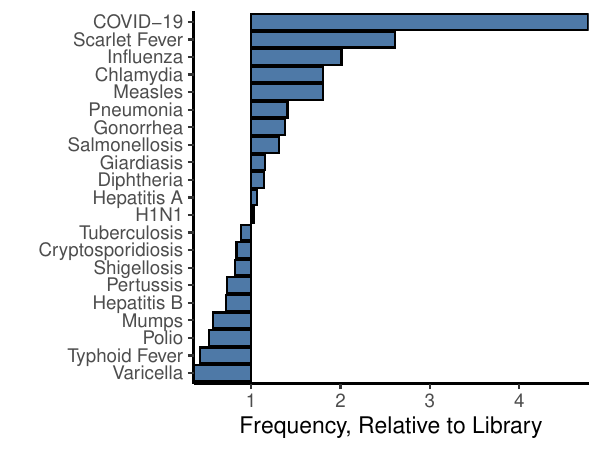}}\\
\subfloat[Neighborhood composition for the US over time ]{\includegraphics[width=0.9\linewidth]{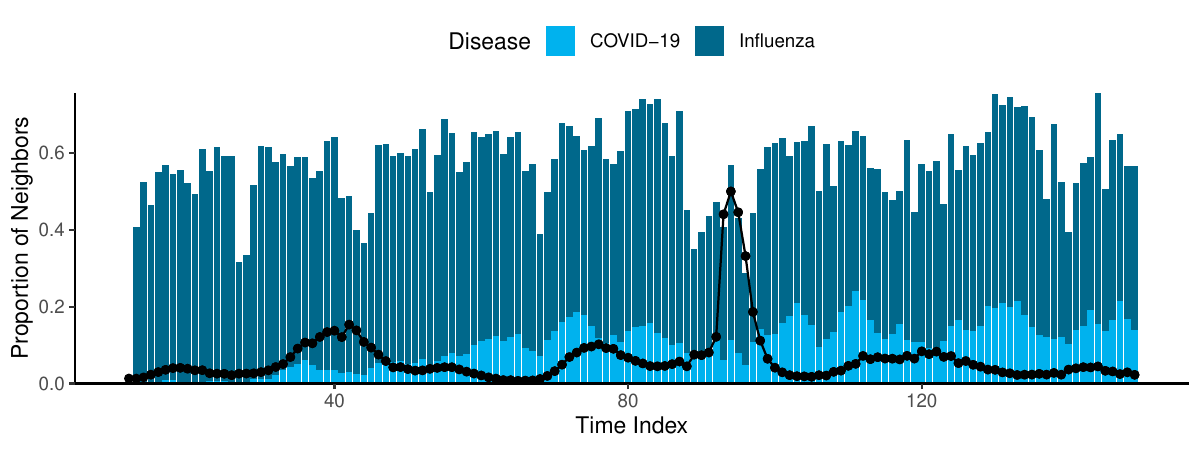}}
\end{figure}

The results of the previous section suggest that including data from other diseases improves COVID-19 forecasts (especially other respiratory diseases). In this section, we examine the forecasts for COVID-19 that include all available data and determine which diseases contribute the most to the forecasts created by the MOA model. The MOA model is useful for determining what diseases are contributing to forecasts since we can examine the identified snippet ``neighborhoods" and record from which diseases these snippets came.  For the forecasts performed on the COVID-19 data, looking at the closest snippets informs which data from the {\color{red}76} other disease data streams ``looked the most like" the COVID-19 data observed up through a given forecast horizon.

We examine the total frequency that each disease's snippets were included in the MOA snippet neighborhoods in \textbf{Figure \ref{fig:covid_casestudy}a}. These frequencies were summed across COVID-19 forecast dates within location and across locations to obtain an overall frequency distribution.  By far, the most frequently used snippets corresponded to influenza data. Of all the chosen snippets, {\color{red}75}\% were from respiratory diseases, where respiratory disease snippets made up {\color{red}63}\% of the available snippets. 

While influenza was included in the snippet neighborhoods the most frequently, there are also more of these data snippets available than any of the other diseases, especially early on in the COVID-19 pandemic.  In \textbf{Figure \ref{fig:covid_casestudy}b}, we scaled these neighborhood composition frequencies relative to the total number of times snippets of each disease were included in the MOA library (again summed across all forecast dates and locations). While the MOA neighborhood algorithm provides no information about the disease label attached to each snippet in the snippet library, it disproportionately includes other COVID-19 snippets in the distance-based neighborhoods (at a rate of over 4x). Perhaps surprisingly, scarlet fever snippets are also disproportionately included in the identified snippet neighborhoods relative to their frequencies in the snippet library overall -- even higher than influenza if you account for the number of influenza snippets available in the snippet library. 

We suspected that within the MOA model the data from onset of the COVID-19 pandemic would become more relevant as they become available, perhaps even replacing influenza in importance towards the end of the pandemic.  We assess this in \textbf{Figure \ref{fig:covid_casestudy}c}, where we color the cumulative proportion of snippets from influenza and COVID-19 over the course of the pandemic for US overall cases. The figure overall supports our suspicion: the proportion of neighbors that come from COVID-19 data increases over the course of the pandemic as more and more COVID-19 snippets become available in the snippet library. However, influenza continues to be heavily used in the MOA forecast throughout the entire pandemic.




\section{Discussion}\label{sec:discussion}

By assembling a uniquely large and diverse dataset spanning {\color{red}66} diseases ({\color{red}76} including subtypes and variations) across {\color{red}13} data sources, we show that modern machine learning models are capable of drawing useful connections across pathogens, modes of transmission, and reporting systems. In doing so, we provide evidence that cross-disease and cross-stream learning can yield more robust forecasts without consistently incurring the risks of negative transfer.  We discovered that this was especially true in the case of COVID-19 forecasting, where adding data from other respiratory diseases showed a marked improvement in forecasting accuracy and uncertainty quantification.

We have shown that expanding the training data has varying effects based on the class of model. The LSTM model benefited most strongly from the inclusion of cross-disease data, reflecting its capacity to learn both short- and long-term temporal dependencies. The GBM model also improved meaningfully when adding data beyond those from a single disease and data source. In contrast, the MOA model benefitted less from more diverse data libraries (likely due to its focus on local time series behavior only), though MOA's interpretability was valuable for identifying which diseases most informed particular forecasts. This interpretability of the MOA model allowed us to analyze what diseases best inform early COVID-19 forecasts, where inclusion of other respiratory diseases such as influenza substantially improved forecast performance. 

A notable outlier in forecast model performance was dengue fever, where inclusion of other vector-borne disease data substantially degraded forecast performance, especially for LSTM and MOA. Our explorations indicated this was likely due to a substantial mismatch between features of the OpenDengue dengue fever data and the other vector-borne diseases, especially in terms of local signal to noise ratio, which was much higher for the other vector-borne diseases. This proves a concrete example of negative transfer, where including additional data results in worse forecast performance. In practice, we would recommend evaluating the data you want to include to determine whether it is demonstrably and systematically different from the data you want to forecast, such as in terms of signal to noise ratio and sample entropy, or in noticeable visual assessments of qualities like seasonality. If the candidate data is very different from the test case, caution should be used including these data unless (i) you have a forecast method that can identify and account for differences between the training and evaluation data or (ii) you have extremely limited data available for the evaluation data stream, in which case incorporating low-quality or very different data could potentially still aid in forecast efforts. 
 
Beyond model performance, the findings in this paper have broader implications for the practice of infectious disease forecasting. Forecasting systems that rely narrowly on a single stream or pathogen will remain brittle, particularly in the face of emerging outbreaks. Pooling wide-ranging historical data provides a pathway toward more resilient, adaptable models that are capable of operating in data-sparse settings. This being said, domain relevance should still be a concern for practitioners. Future research should focus on principled approaches to selecting and screening auxiliary data sources, identifying when additional information is likely to improve forecasts and when it risks adding noise or causing negative transfer. 

Finally, this work contributes not only methodological advances but also a substantial data resource. The harmonized, cleaned repository of disease time series is publicly available at \url{https://github.com/lanl/precog/tree/main/infectious_timeseries_repo}, enabling future studies to fine-tune modern ML models for diverse forecasting tasks. See individual data stream READMEs in that repository for detailed licensing information. 

This paper shows that expanding the scope of training data beyond the data stream being forecasted is both feasible and, in the majority of cases, beneficial. As forecasting becomes an increasingly central component of public health preparedness, cross-disease and cross-data stream transfer learning offers a practical and scalable strategy for improving the accuracy, stability, and robustness of predictive models, especially in data-sparse scenarios.

\section*{Acknowledgments}
Research presented in this article was supported by the Laboratory Directed Research and Development program of Los Alamos National Laboratory under project number 20240066DR. Los Alamos National Laboratory is operated by Triad National Security, LLC, for the National Nuclear Security Administration of U.S. Department of Energy (Contract No. 89233218CNA000001). This work has been approved for public release under LA-UR-26-24105.

\section*{Code and Data Availability}
All data used in this work have been made publicly available at \url{https://github.com/lanl/precog/tree/main/infectious_timeseries_repo}. This repository also contains additional data not included in this work, including monthly data collected from the US CDC Foodborne Diseases Active Surveillance Network (Foodnet) \citep{Simpson2020}. Code for this work is available at \url{https://github.com/lanl/precog/tree/main/crossdisease}. 

\bibliographystyle{chicago}
\bibliography{sample}

\end{document}